\documentclass[10pt,twocolumn,letterpaper]{article}

\usepackage{iccv}
\usepackage{times}
\usepackage{epsfig}
\usepackage{graphicx}
\usepackage{amsmath}
\usepackage{amssymb}

\usepackage{accsupp}[axessibility]  
\usepackage{multirow}
\usepackage{graphicx}
\usepackage{enumitem}
\usepackage{rotating}
\usepackage{lipsum}
\usepackage{tabularx}
\usepackage{epstopdf}
\usepackage{makecell}

\usepackage[breaklinks=true,bookmarks=false]{hyperref}

\iccvfinalcopy 

\def\iccvPaperID{****} 

\ificcvfinal\fi

\begin{document}

\title{PointDC: Unsupervised Semantic Segmentation of 3D Point Clouds via Cross-modal Distillation and Super-Voxel Clustering}

\author{
Zisheng Chen\textsuperscript{\rm 1}\thanks{These authors contributes equally.},
Hongbin Xu\textsuperscript{\rm 1,\rm 2}\footnotemark[1],
Weitao Chen\textsuperscript{\rm 2},
Zhipeng Zhou\textsuperscript{\rm 3},
Haihong Xiao\textsuperscript{\rm 1},
Baigui Sun\textsuperscript{\rm 2},\\
Xuansong Xie\textsuperscript{\rm 2},
Wenxiong kang\textsuperscript{\rm 1,\rm 4}\thanks{Corresponding author.}\\
\textsuperscript{\rm 1}South China University of Technology\\
\textsuperscript{\rm 2}Alibaba Group,
\textsuperscript{\rm 3}Chinese Academy of Science,
\textsuperscript{\rm 4}Pazhou Laboratory\\
{\tt\small halveschen@163.com}\quad{\tt\small {hongbinxu1013,hillskyxm}@gmail.com}\quad{\tt\small auwxkang@scut.edu.cn}
}


\maketitle
\ificcvfinal\thispagestyle{empty}\fi

\begin{abstract}
Semantic segmentation of point clouds usually requires exhausting efforts of human annotations, hence it attracts wide attention to the challenging topic of learning from unlabeled or weaker forms of annotations.
In this paper, we take the first attempt for fully unsupervised semantic segmentation of point clouds, which aims to delineate semantically meaningful objects without any form of annotations.
Previous works of unsupervised pipeline on 2D images fails in this task of point clouds, due to: 1) Clustering Ambiguity caused by limited magnitude of data and imbalanced class distribution; 2) Irregularity Ambiguity caused by the irregular sparsity of point cloud.
Therefore, we propose a novel framework, PointDC, which is comprised of two steps that handle the aforementioned problems respectively: Cross-Modal Distillation (CMD) and Super-Voxel Clustering (SVC).
In the first stage of CMD, multi-view visual features are back-projected to the 3D space and aggregated to a unified point feature to distill the training of the point representation.
In the second stage of SVC, the point features are aggregated to super-voxels and then fed to the iterative clustering process for excavating semantic classes.
PointDC\footnote{The code is released at: \url{https://github.com/SCUT-BIP-Lab/PointDC}.} yields a significant improvement over the prior state-of-the-art unsupervised methods, on both the ScanNet-v2 (+18.4 mIoU) and S3DIS (+11.5 mIoU) semantic segmentation benchmarks.
\end{abstract}

\section{Introduction}
\label{sec:introduction}


Semantic segmentation of 3D point cloud is a crucial problem that assigns each individual point to a known ontology.
The segmentation models can delineate objects at a fine granularity of object boundary, which is helpful for multiple downstream applications, such as robotic navigation, autonomous vehicles, and scene parsing. 
Despite the immense progress of fully-supervised schemes in 3D semantic segmentation, the success crucially relies on large-scale datasets and annotations.
Unfortunately, it requires exhausting efforts to conduct semantic-level per-point annotations (e.g. $\approx 22.3$ minutes per indoor scene for annotation \cite{scannet}).

\begin{figure}[t]
  \centering
  \includegraphics[width=0.45\textwidth]{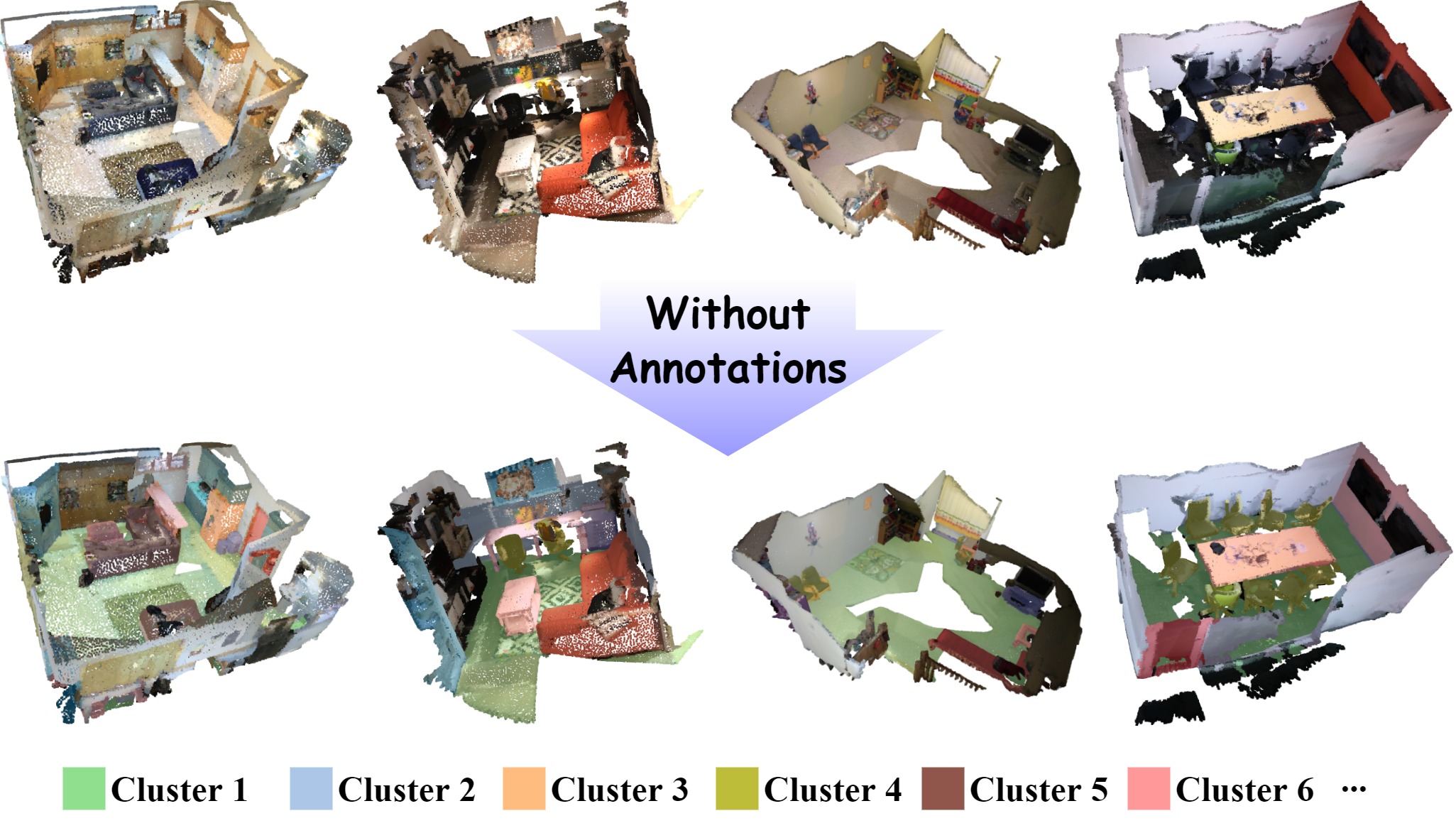}
  \caption{From unannotated point clouds, we would like a segmentation system to discover the semantic concepts automatically without any supervision.}
  \label{fig:motivation}
\end{figure}

To reduce the efforts on the tedious process of semantic annotations, several works were proposed to create 3D semantic segmentation systems that can be trained from weaker forms of annotations, including projected 2D images \cite{wang2020weakly}, subcloud-level \cite{wei2020multi}, segment-level \cite{tao2022seggroup}, and point-level annotations \cite{liu2021one, wu2022pointmatch, hu2022sqn}. 
However, few works attempt to handle the great challenge of 3D semantic segmentation without any form of human annotations or motion cues.
In this paper, we aim to build an unsupervised 3D semantic segmentation framework that can automatically excavate meaningful semantic features from point clouds, as shown in Fig. \ref{fig:motivation}.

Some efforts of unsupervised semantic segmentation have been witnessed in the research field of 2D visual images.
Independent Information Clustering (IIC) \cite{iic} and PiCIE \cite{picie} try to excavate semantically meaningful features through self-supervised learning based on transformation invariance and equivariance, meantime conducting a clustering process to optimize the compactness of the learned semantic clusters.
STEGO \cite{STEGO} further distills from a self-supervisedly pretrained Transformer to form compact semantic clusters of the corpora.
Whereas the existing line of unsupervised clustering pipeline could not be simply migrated to 3D point clouds for the following reasons:

1) \textbf{Clustering Ambiguity}: 
For a 2D unsupervised system, the major premise is that the pictures are meaningful images collected by human (rather than meaningless case like an empty image with a single color).
The nature prior during the collection of human enables the effectiveness of large-scale unsupervised clustering on 2D images as long as the dataset is large enough.
However, for a 3D unsupervised system, on the one hand, the huge cost for data collection limits the magnitude and diversity of the dataset; on the other hand, the imbalanced occupancy of 3D space aggravates the long-tail distribution effect among different classes when clustering among points, resulting in the ignorance of classes with fewer points.

2) \textbf{Irregularity Ambiguity}:
Without a regular grid-like structure, point clouds might have variations in the density of local areas.
During the calculation of clusters, the points from dense areas inherently weigh more than the points from sparse areas.
As a result, the original K-means clustering in \cite{iic,picie} might overly focus on the dense regions and ignores the sparse regions.

In this paper, we take the first attempt for unsupervised 3D semantic segmentation and introduce \textbf{PointDC} (\textbf{Point} cloud cross-modal \textbf{D}istillation and Super-Voxel \textbf{C}lustering), which is capable of discovering and segmenting semantic objects from point clouds without any human annotations.
Directing to handle the aforementioned problems of Clustering Ambiguity and Irregularity Ambiguity, we respectively adopt Cross-Modal Distillation (CMD) and Super-Voxel Clustering (SVC) in our PointDC framework.

1) To handle the former problem, CMD could integrate the multi-view visual clues to distill the corresponding point feature in the point cloud.
As cognitive scientists \cite{deloache1979picture,rose1977infants} argue,  humans are proficient at mapping the visual concepts learned from 2D images to understand the 3D world.
We obtain multi-view images by observing 3D point clouds from different viewpoints first and feed them to a self-supervisedly pretrained visual model such as DINO \cite{dino} to extract unsupervised visual features.
By back-projecting the multi-view visual features into the corresponding points in 3D space, we can aggregate the features from different views to formulate a unified multi-view representation, and distill the learning of point representation.
The involvement of multi-view visual cues can effectively diminish the ambiguity during clustering among points, and provide a coarse understanding of 3D semantic features.

2) To handle the latter problem, instead of clustering on the original point space, SVC rasterizes the 3D space into super-voxels and assigns each point to the corresponding super-voxel.
The features of points in the same voxel are aggregated together via Super-Voxel Pooling for a unified permutation-invariant representation.
During each iteration phase of clustering process, we first assign the point features to the super-voxels and then cluster among these voxels.
Afterward, the feature of each super-voxel is assigned back to the occupying points, and assume these points in the local super-voxel share a common semantic feature.

The overall pipeline of our PointDC framework includes 2 steps: 1) CMD is utilized to distill the learning of point representation first; 2) Then SVC is applied iteratively to optimize the clustered semantic representation on point clouds.
For evaluation, we conduct extensive experiments on the challenging ScanNet-v2 \cite{scannet} and S3DIS \cite{s3dis}.
Compared with state-of-the-art on existing unsupervised methods for point cloud semantic segmentation, our PointDC achieves an improvement on both the ScanNet-v2 (\textbf{+18.4 mIOU}) and S3DIS (\textbf{+11.5 mIOU}).

In summary, our contribution is threefold.
\begin{itemize}[itemsep=2pt,topsep=0pt,parsep=0pt]
	\item We take the first attempt for unsupervised 3D semantic segmentation without any kinds of human annotations.
	\item We propose PointDC, a novel framework for unsupervised 3D semantic segmentation.
    It is comprised of 2 steps: 1) Cross-Modal Distillation that distills multi-view visual features to the point-based representations; 2) Super-Voxel Clustering that regularizes the point features with voxelized representation through super-voxel pooling, and iteratively clusters to optimize the semantic features on point clouds.
	\item The proposed method achieves superior improvement compared with existing unsupervised methods for 3D semantic segmentation on various challenging datasets, demonstrating its effectiveness.
\end{itemize}

\section{Related Work}

\begin{figure*}[t]
	\centering
	\includegraphics[width=0.9\textwidth]{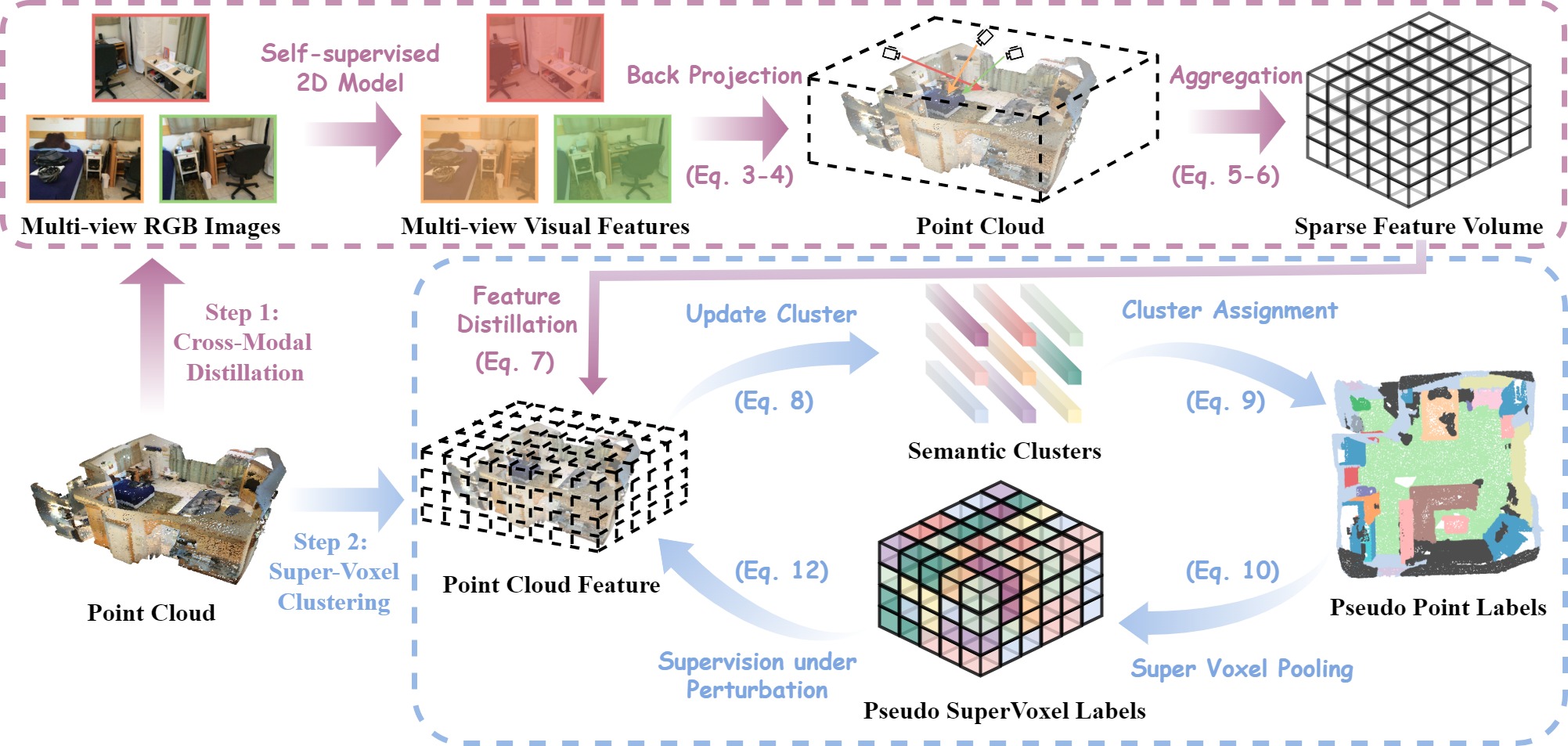}
	\caption{Overview of PointDC framework.
                The training contains 2 steps: Cross-Modal Distillation and Super-Voxel Clustering.}
	\label{fig:framework}
\end{figure*}

\noindent\textbf{3D Semantic Segmentation} 
3D semantic segmentation approaches can be divided into 2 categories: point-based methods \cite{pointnet, pointnet++, dgcnn, pointcnn,superpointcloudv1, superpointcloudv2} and voxel-based methods \cite{sparseconv,mkconv}.
In point-based methods \cite{dgcnn, pointcnn,superpointcloudv1, superpointcloudv2}, the information of points is fused from their neighboring areas computed from K-NN or spherical search for effective 3D representations.
In voxel-based methods, the points in the 3D space are converted to voxels with 3D-grid structure.
In these voxelized representations, standard convolution operations can be applied to extract features from 3D information.
Due to the sparsity of point clouds, sparse convolution \cite{sparseconv, mkconv} is adopted to process the voxelized representation of point clouds.
Recently, the Transformer structure \cite{zhao2021point} is also used to handle point clouds, as a novel alternative to the classic convolutional structure.
However, most of the previous works are designed for fully-supervised schemes or utilizing weaker forms of annotations \cite{wu2022pointmatch}.
In this work, we focus on handling the 3D semantic segmentation problem without using any human annotations.

\noindent\textbf{2D Unsupervised Learning} 
There has been a number of recent progresses in unsupervised 2D semantic segmentation.
DeepCluster \cite{deepcluster} obtains latent semantic representations by clustering in the low-dimensional feature space.
IIC \cite{iic} proposes invariant-information clustering in pixel-level representation to learn by clustering with a self-supervised manner.
PiCIE \cite{picie} leverages the inductive bias of photometric invariance and geometric equivariance, bringing two sets of images with different transformations close to each other in the feature space.
MaskContrast \cite{maskcontrast} uses an unsupervised saliency detection model to obtain a binary encoding of the silhouette of salient objects in the image, and further adopts contrastive learning to pull the distances of pixel-wise features inside the mask and push the ones between different masks.
STEGO \cite{STEGO} extracts features from pretrained models and proposes a novel contrastive loss that encourages features to form compact clusters while preserving the relationships across the corpora.


\noindent\textbf{3D Unsupervised Learning} 
3D unsupervised learning can be roughly divided into two categories: generation-based methods\cite{occo, PU-GAN} and contrastive-learning-based methods\cite{pointcontrast, csc, strl}.
Generation-based methods let the model to complete the input point clouds which are occluded or down-sampled.
Contrastive-learning-based methods define the augmentation (rotation, color jittering, different views) of the given point clouds  as positive samples and others as negative samples. 

\noindent\textbf{Cross-Modal Learning} 
Cross-Modal learning utilizes data from different modalities\cite{crosspoint, image2point, learningfrom2d}. 
\cite{image2point} directly converts a 2D pretrained model into a point cloud model using filter dilation. 
\cite{crosspoint} defines images as strong positive samples, then enforces both intra-modal and cross-modal global feature correspondence in the invariant space.

\section{Method}

In this work, we first attempt in learning unsupervised 3D semantic segmentation models only given uncurated and unlabeled datasets of point clouds.
This section begins by introducing the problem statement (Sec. \ref{sec:method:problem}).
We formulate this task as point-level clustering among the whole point cloud dataset to discover semantic meaningful clusters, and introduce the preliminary of learning by clustering pipelines in Sec. \ref{sec:method:prelinminary}.
To handle the problem of Clustering Ambiguity and Irregularity Ambiguity in existing clustering pipelines, we propose Cross-Modal Distillation (CMD) (Sec. \ref{sec:method:cmd}) and Super-Voxel Clustering (SVC) (Sec. \ref{sec:method:svc}).
Finally, the overall training process of our PointDC framework is introduced in Sec. \ref{sec:method:overall}.
The overview of the PointDC is shown in Fig. \ref{fig:framework}.

\subsection{Problem Statement}
\label{sec:method:problem}

Given the unlabeled dataset $\mathcal{U}$ from some domain $\mathcal{D}$ which has $M$ point clouds in total.
The $i$-th point cloud ${P}_i \in \mathbb{R}^{N \times 6}$ represents a scene with $N$ points of 6 dimensions including the XYZ coordinates and RGB intensities of points.
On this dataset $\mathcal{U}$, we aim to discover a set of virtual semantically meaningful classes $\mathcal{C}$ and learn a semantic feature extractor $f_\theta$ parameterized by $\theta$.
When provided an unseen point cloud from domain $\mathcal{D}$ during evaluation, $f_\theta$ should be able to assign every point a label from the discovered classes $\mathcal{C}$.

\subsection{Preliminary}
\label{sec:method:prelinminary}

We begin with preliminaries of prior works that learn an end-to-end neural network for clustering unlabeled data \cite{iic,picie,STEGO,mei2022unsupervised,zhang2019unsupervised}.
The key point in these works is that clustering data into classes requires strong feature representation, meantime the learning of feature representations also needs precise class labels.
To handle this chicken-and-egg problem, the simplest solution is the one defined by DeepCluster \cite{tian2017deepcluster}.
Following the procedure of an E-M algorithm, we can alternate between clustering using currently extracted feature representation, and using the clustered results as pseudo-labels to supervise the training of the feature extractor.
We can still follow this simple strategy for point cloud semantic segmentation task, by alternating the clustering process among instance-wise feature to point-wise feature representation.

Concretely, suppose that we have a set of unlabeled point clouds $P_i$, where $i$ represents the index in the dataset.
The extracted feature tensor is denoted as $f_{\theta}(P_i) \in \mathbb{R}^{N \times D}$, where $D$ is the feature dimension.
Denote $f_{\theta}({P}_i)[j]$ as the feature vector on the $j$-th point of point cloud $P_i$ and $\mu \in \mathbb{R}^{C \times D}$ as the randomly initialized cluster centroids.
The baseline of learning by clustering can be summarized as follows:

\begin{enumerate}[itemsep=2pt,topsep=0pt,parsep=0pt]
	\item Optimizing the following object function and use K-Means to cluster the current feature among all points in the dataset:
	\begin{equation}
		\min_{y,\mu} \sum_{i,j} \| f_{\theta}({P}_i)[j] - \mu[y_{ij}] \|
	\end{equation}
        where $y_{ij}$ is the assigned label of the $j$-th point of $i$-th point cloud.
	\item Use the clustered labels as pseudo-labels to train the segmentation network:
	\begin{equation}
		\min_{\theta,\omega} \sum_{i,j} L_{CE} (g_{\omega} ( f_{\theta} (P_i) [j] ), y_{ij} )
		\label{eq2}
	\end{equation}
	where $L_{CE}$ is the cross-entropy function and $g_{\omega}$ is the segmentation head parameterized by $\omega$.
\end{enumerate}

\subsection{Cross-Modal Distillation}
\label{sec:method:cmd}

\begin{figure}[t]
  \centering
  \includegraphics[width=0.4\textwidth]{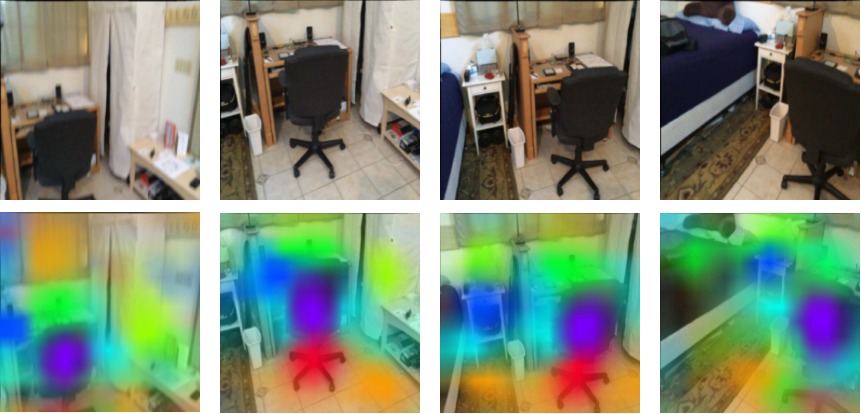}
  \caption{Visualization of the clustering results among multi-view feature maps extracted by DINO \cite{dino}. It demonstrates that the multi-view features are semantically correlated.}
  \label{fig:multiview_clustering}
\end{figure}

As discussed in Sec. \ref{sec:introduction}, the Clustering Ambiguity problem in point cloud is caused by the extremely imbalanced occupancy of different classes in 3D space and the lack of large-scale point cloud datasets with diversity and magnitude comparable with image datasets.
Instead of directly clustering on the point cloud dataset, we propose Cross-Modal Distillation (CMD) as an initialization step before clustering.
In intuition, the multi-view visual features are semantically correlated coarsely, as shown in Fig. \ref{fig:multiview_clustering}.
Hence, the distillation from multi-view visual modality to 3D point cloud modality can provide a reliable initialization for clustering.


In CMD, the point cloud $P_i$ is converted to images $I_{iv}$ by observing $P_i$ on different viewpoints, where $v$ is the index of viewpoints.
Given a self-supervised pretrained 2D neural network $h$, we can obtain the visual feature map $h(I_{iv})$ of each image $I_{iv}$.
Suppose that the intrinsic and extrinsic matrix of camera $k$ is respectively $K_v$ and $T_v$.
By projecting the 3D point cloud $P_i$ to each of the multi-view images, we can calculate the cross-modal correspondence between the 3D points and 2D pixels.

Given the $j$-th point of $P_i$, we can calculate its projection $\hat{p}_{ijv}$ on view $v$ via:
\begin{equation}
	z_{ijv} \hat{p}_{ijv} = K_v T_v P_i[j]
	\label{eq3}
\end{equation}
where $z_{ijv}$ is the depth value on $\hat{p}_{ijv} = [u_{ijv}, v_{ijv}, 1]^T$. $u_{ijv}$ and $v_{ijv}$ are the image pixel coordinates on width and height.

Then, we can filter the invalid projections which are outside the imaging areas with $u_{ijv} > W$, $u_{ijv} < 0$, $v_{ijv} > H$, and $v_{ijv} < 0$.
Denote that the remaining projections as $\tilde{p}_{ijv}$, and its depth value as $\tilde{z}_{ijv}$.
Since there might exist multiple projections to the same pixel, we filter the occluded points by finding the projection with minimum depth value:
\begin{equation}
	j^* = \mathop{\arg\min}\limits_{j} \tilde{z}_{ijv}
	\label{eq4}
\end{equation}
where $j^*$ represents the index of the corresponding point on point cloud $P_i$.

Afterward, we can warp the pixel-wise feature to the corresponding point indexed by $j^*$:
\begin{equation}
	H_{iv} = h(I_{iv})[\tilde{p}_{ij^*v}]
	\label{eq5}
\end{equation}
where $H_{iv} \in \mathbb{R}^{N \times V \times D}$ is the point-wise feature projected from visual features on multiple views.

Suppose that the super-voxel segmentation function is $S_{vox}( \cdot )$.
The feature $F_i \in \mathbb{R}^{M \times D}$ on the super-voxel $S_{vox}(P_i) \in \mathbb{R}^{M \times 3}$ converted from original point cloud $P_i$ can be calculated through:
\begin{equation}
	F_i[k] = \frac{1}{\|\mathcal{N}(k)\|}\sum_{j \in \mathcal{N}(k)} \max_{v} H_{iv}[j]
	\label{eq6}
\end{equation}
where $k$ is the index of super-voxel and $\mathcal{N}(k)$ represents the indices of points belonging to the $k$-th super-voxel.
The multi-view features are firstly aggregated via Max-Pooling among different views and then the feature on each super-voxel is aggregated via Avg-Pooling among the occupying points.

Finally, in the stage of CMD, the point feature extractor is then supervised by the visual feature aggregated from multi-view images.
CMD optimizes the following function while traversing the whole dataset:
\begin{equation}
	\min_{\theta} \sum_{i} \| \left( \frac{1}{\|\mathcal{N}(k)\|}\sum_{j \in \mathcal{N}(k)} (f_{\theta}(P_i)[j]) \right) - F_i \|_2^2
	\label{eq7}
\end{equation}
where we first aggregate the point feature $f_{\theta}(P_i)$ in each super-voxel via AVG-Pooling and distill it with $F_i$.

\subsection{Super-Voxel Clustering}
\label{sec:method:svc}

As discussed in Sec. \ref{sec:introduction}, the Irregularity Ambiguity problem comes from the irregular structure of point cloud.
In a scene, the dense points might disturb the distance-based clustering metric, leading to the ignorance of sparse points.
To handle this issue, we propose Super-Voxel Clustering (SVC), an iterative learning by clustering pipeline that alternates point-based clustering with super-voxel-based clustering.


In SVC, given the point cloud $P_i \in \mathbb{R}^{N \times 3}$.
Denote that the feature extractor $f_{\theta} (\cdot)$ is parameterized by $\theta$.
The number of clusters is $C$, the output feature dimension of $f_{\theta} ( \cdot)$ is $D$, and the number of super-voxels is $M$. 
The learning process can be summarized as follows:
\begin{enumerate}[itemsep=2pt,topsep=0pt,parsep=0pt]
	\item Extract point-wise feature from point cloud $P_i$ and update the super-voxelized features via K-Means:
	\begin{equation}
		\footnotesize
		y, \mu^* = \mathop{\arg\min}_{y,\mu} \sum_{i,k} \| S_{vox}(f_\theta(P_i)) [k] - \mu[y_{ik}] \|_2^2
		\label{eq8}
	\end{equation}
	where $i$ is the index of point cloud in the dataset and $k$ is the index of super-voxel.
	The super-voxel aggregation function $S_{vox}(\cdot)$ aggregates the feature of points belonging to same super-voxel: $S_{vox} ( f_\theta(P_i) ) =  \frac{1}{\|\mathcal{N}(k)\|}\sum_{j \in \mathcal{N}(k)} f_\theta(P_i)[j]$.
	$y \in \mathbb{R}^{M \times C}$ is the assigned label on super-voxels, and $\mu^* \in \mathbb{R}^{C \times D}$ is the clustered centroids.

	\item To assign the label to each point of the point cloud, we use the distance to the clustered centroids $\mu^*$ to create soft assignment of probabilities towards each class.
	\begin{equation}
	\hat{y}_{ij} = -\log \left(\frac{e^{-\text{cos}(f_\theta (P_i)[j], \mu^*[\hat{y}_{ij}])}}{\sum_l e^{-\text{cos}(f_\theta (P_i)[j], \mu^*_l)}} \right)
		\label{eq9}
	\end{equation}
	where $\hat{y}_{ij}$ is the assigned point-wise pseudo labels on the $j$-th point of point cloud $P_i$.
	$\text{cos} (\cdot, \cdot)$ is the cosine distance.
	
	\item  Then, we adopt Super-Voxel Pooling to filter the assigned point-wise labels in each voxel of the super-voxel.
	\begin{equation}
		\tilde{y}_{ik} = \frac{1}{\|{N}(k)\|}\sum_{j \in \mathcal{N}(k)} \hat{y}_{ij}
		\label{eq10}
	\end{equation}
	where $\tilde{y}_{ik}$ is the soft label filtered by the prior of super-voxels and $k$ is the index of super-voxel.

	\item Finally, we can train the point-wise representation with  $\tilde{y}_{ik}$.
	\begin{equation}
		\theta^* = \mathop{\arg\min}_{\theta} \| \sum_{k} \sum_{j \in \mathcal{N}(k)} L_{CE} (f_{\theta} (P_i)[j] \otimes \mu^*, \phi(\tilde{y}_{ik})) \|^2_2
		\label{eq11}	
	\end{equation}
	where $L_{CE}$ is the cross-entropy function, $\otimes$ represents the calculation of similarity between point feature and clustered centroids, $\phi (\cdot) = \text{onehot} ( \mathop{\arg\max} (\cdot))$ converts the input an one-hot label.
	
	\item Repeat previous 1-4 operations iteratively.
	
\end{enumerate}

Furthermore, the inductive bias of invariance and equivariance to specific transformations are used to promote the pseudo-label training in Eq. \ref{eq11}.
Following the self-supervised prior, the extracted features should maintain the invariance towards transformations like color jittering or gaussian noise, and the extracted feature should be warped or rotated accordingly to the geometric transformation on the original point cloud to keep the equivariance.
Denote that $\pi_{inv} (\cdot)$ and $\pi_{equ} (\cdot)$ are respectively the transformation for invariance and equivariance.
Then the inductive bias could be appended to promote Eq. \ref{eq11}:
\begin{equation}
	\footnotesize
	\theta^* = \mathop{\arg\min}_{\theta} \| \sum_{k} \sum_{j \in \mathcal{N}(k)} L_{CE} (f_{\theta} (\pi_{equ}(\pi_{inv}(P_i)))[j], \phi(\pi_{equ}(\tilde{y}_{ik}))) \|^2_2
	\label{eq12}
\end{equation}

\subsection{Overall Framework}
\label{sec:method:overall}

As presented in Fig. \ref{fig:framework}, our PointDC framework includes two training stages: CMD and SVC.
In the first stage of CMD, we can obtain multi-view images by observing the point cloud from different viewpoints.
A self-supervisedly pretrained 2D model is used to extract the feature maps from multi-view images.
Then we can back-project (Eq. \ref{eq5}) each pixel of the multi-view images to its corresponding point of the point cloud by calculating Eq. \ref{eq3} and \ref{eq4}.
Since one point might have multiple projections on different images, we aggregate the cross-view feature via the Global Max-Pooling in Eq. \ref{eq6}.
Following the assumption that each super-voxel contains similar semantics, we further aggregate the features of points belonging to the same super-voxel via the Global Avg-Pooling in Eq. \ref{eq6}.
The feature extractor is then distilled by the multi-view cues in Eq. \ref{eq7}.
In the second stage of SVC, we perform K-Means clustering on the super-voxels aggregated from point-wise features in local regions in Eq. \ref{eq8}.
Then we assign the label to each point of the point cloud in a non-parametric manner based on the distance to clusters in Eq. \ref{eq9}.
The label for each super-voxel is then aggregated via Super-Voxel Pooling among points located in the same voxel in Eq. \ref{eq10}.
Finally, the pseudo label of super-voxel is used to train the feature extractor under random perturbation of transformations in Eq. \ref{eq12}.

%
%

\section{Experiments}

\begin{table}[t]
\centering
\resizebox{0.8\hsize}{!}{
\begin{tabular}{l|cc|cc}
\hline
\multirow{2}{*}{{Methods}} & \multicolumn{2}{l|}{{Unsupervised}} & \multicolumn{2}{l}{{Linear Probe}} \\
                         & {mIoU}             & {Acc}            & {mIoU}            & {Acc}            \\ \hline \hline
 \cite{deepcluster} DeepCluster  & 3.88 & 19.75 & 4.04 & 23.55 \\    
 \cite{iic} IIC & 3.98 & 20.47 & 4.00 & 23.06  \\
 \cite{picie} PiCIE & 4.10 & 22.81 & 4.34 & 27.04  \\ \hline
 \cite{pointcontrast} PC-HC & 4.63 & 21.75 & 4.72 & 43.09   \\
 \cite{pointcontrast} PC-NCE & 3.93 & 21.24 & 4.29 & 44.21  \\
 \cite{occo} OcCo & 3.17 & 19.97 & 3.37 & 21.65 \\
 \cite{crosspoint} CrossPoint & 3.81 & 20.52 & 3.94 & 22.92 \\
 \cite{csc} CSC & 4.64 &  18.24  & 5.31 & 28.5   \\
 \cite{strl} STRL & 4.13 & 19.38 & 4.25 & 29.70 \\ \hline
 PointDC & \textbf{25.74} & \textbf{63.69}  & \textbf{28.78} & \textbf{71.62}  \\ \hline
\end{tabular}
}
\caption{Comparison of unsupervised segmentation on the ScanNet-v2 validation set. PointDC significantly outperforms prior art in both unsupervised clustering and linear probe metrics.}
\label{tab:scannet_val}
\end{table}

\begin{table}[t] 
\centering
\resizebox{0.42\hsize}{!}{
\begin{tabular}{l| c c } \hline
    Method & mIoU \\
    \hline \hline
    \cite{deepcluster} DeepCluster & 3.7  \\
    \cite{iic} IIC & 3.7  \\
    \cite{picie} PiCIE &  3.9  \\
    \hline
    \cite{pointcontrast} PC-HC &  3.9  \\
    \cite{pointcontrast} PC-NCE & 3.8  \\
    \cite{csc} CSC & 4.5\\
    \cite{strl} STRL & 4.1 \\
    \hline
    PointDC & \textbf{22.9}   \\
    \hline
\end{tabular}
}
\caption{Comparison of unsupervised segmentation on the ScanNet-v2 test set. The results of the online benchmark are reported.}
\label{tab:scannet_test}
\end{table}

\subsection{Experiment Details} 
\label{Implement details}

\noindent\textbf{Datasets and Metric:} 
We conduct experiments on 2 point cloud benchmarks, ScanNet-v2\cite{scannet} and S3DIS\cite{s3dis}.
ScanNet-v2\cite{scannet} contains 1613 3D scans from 707 unique indoor scenes, all annotated with 20 classes.
Following the  official setting, we use 1201 scenes and 312 scenes as training set and validation set, respectively.
The remaining 100 scenes are used as test set.
S3DIS\cite{s3dis} contains 271 indoor scenes with 13 classes.
We follow the official train/validation split, training on Areas 1,2,3,4,6 and then testing on Area 5.
As our method requires image data and camera intrinsic and extrinsic parameters, we use ScanNet-v2 2D data and 2D-3D-S\cite{s3dis}.
2D-3D-S contains multi-view images corresponding to the scenes in S3DIS, the corresponding depth maps as well as the internal and external camera parameters.
We utilize the intersection-over-union as evaluation metric of the 3D semantic segmentation results, and report the mean result (mIOU) over all categories for comparison with other approaches.
Moreover, we also utilize the accuracy over all categories in the results.


\begin{table}[t] 
\centering
\resizebox{0.6\hsize}{!}{
\begin{tabular}{l| cc} \hline
    Method & mIoU & Acc \\
    \hline \hline
    \cite{deepcluster} DeepCluster & 5.46 & 19.75 \\
    \cite{iic} IIC & 5.33 & 21.47 \\
    \cite{picie} PiCIE &  5.90 & 25.05 \\
    \hline
    \cite{pointcontrast} PC-HC &  9.27 &  26.87 \\
    \cite{pointcontrast} PC-NCE & 8.86 & 23.32 \\
    \cite{csc} CSC & 11.09 & 34.83 \\
    \cite{strl} STRL & 10.21 & 37.40 \\
    \hline
    PointDC & \textbf{22.59} & \textbf{54.10}  \\
    \hline
\end{tabular}
 }
 \centering
 \caption{Comparison of unsupervised segmentation on the S3DIS validation set (Area 5).}
 \label{tab:s3dis_val}
\end{table}

\begin{figure*}[t]
\centering
\includegraphics[width=0.8\textwidth]{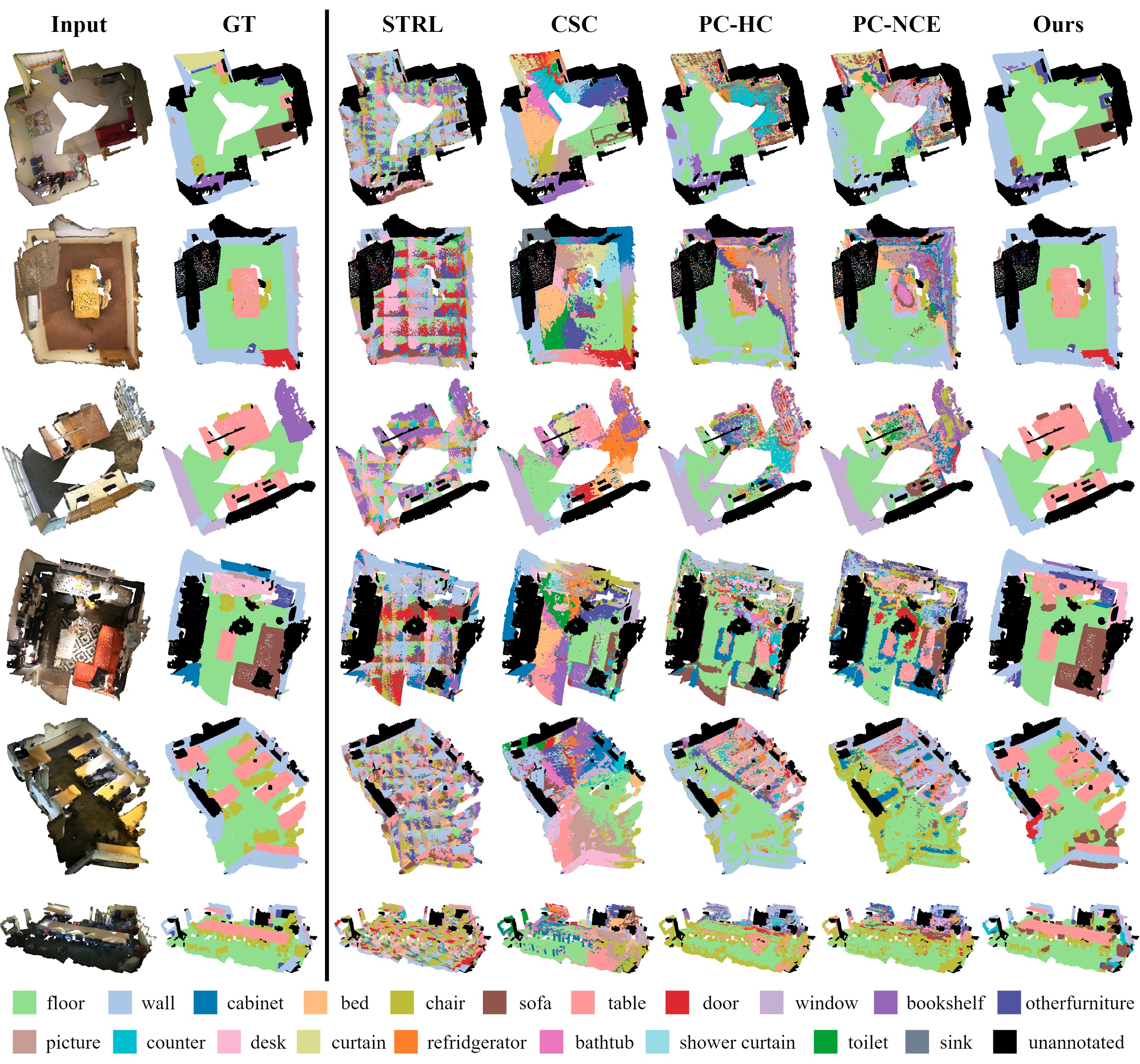}
\caption{Qualitative comparison of unsupervised segmentation on ScanNet-v2 validation set. Each of the aligned ground truth labels and clusters is assigned a color. For better understanding, we show some the color and name matches in the bottom.}
\label{fig:scannet_visualization}
\end{figure*}

\noindent\textbf{Implementation Details:}
In the stage of CMD,  we experiment with pretrained image segmentation model STEGO\cite{STEGO} noted as $h$ in Sec. \ref{sec:method:cmd}.
For the backbone of point feature extractor $f_{\theta}$, we use the Sparse 3D-UNet\cite{3dunet} from \cite{sparseconv}.
The points are voxelized following the procedure of \cite{pointgroup} and then fed to $f_{\theta}$ for feature extraction.
For the super-voxels segmentation \cite{overseg}, we use the mesh segment results \cite{scannet} for ScanNet-v2.
For the super-voxel partition in S3DIS, we utilize the geometric partition results described in \cite{superpointcloudv1}.

\noindent\textbf{Unsupervised Clustering and Linear Probe:}
Since we are agnostic to the ground truth label, the clustered results might have a random permutation of order compared with the ground truth.
To evaluate the quality of an unsupervised method, we follow the 2 protocols used in previous works \cite{maskcontrast, picie}: unsupervised clustering and linear probe.
For unsupervised clustering, we do not have access to the ground truth labels, but we can use a Hungarian matching algorithm to align our unlabeled clusters and the ground truth labels for evaluation.
This measures how consistent the predicted semantic segments are with the ground truth annotations and diminish the aforementioned permutations of the predicted class labels.
For linear probe, we train a linear projection from the features to the class labels with cross-entropy loss.
This measures the feature quality of the learned feature representation.

\begin{table*}[h]
\centering
\resizebox{0.65\linewidth}{!}{
\begin{tabular}{c c c c c c | c c} \hline
    \makecell{Random \\ Clustering} & \makecell{Multi-View \\ Clustering} & \makecell{CMD \\ (Eq. \ref{eq7})} & \makecell{Basic-SVC \\ (Eq. \ref{eq8} and Eq. \ref{eq12})} & \makecell{Non-parametric \\ Classifier (Eq. \ref{eq9})} & \makecell{Super-Voxel \\ Pooling (Eq. \ref{eq10})} & mIoU & Acc \\
    \hline \hline
     \checkmark & & & & & & 3.86 & 13.18  \\
      \checkmark & \checkmark & & & & & 13.58 & 34.00 \\
     \checkmark & \checkmark & \checkmark & & & & {20.29} & 42.12  \\
     \checkmark & \checkmark & \checkmark & \checkmark& & & 23.64  & 61.53 \\
      \checkmark & \checkmark & \checkmark &  \checkmark & \checkmark & & {24.85} & 61.79 \\
      \checkmark & \checkmark & \checkmark &  \checkmark & \checkmark & \checkmark & 25.74 & 63.69 \\
      
    \hline
\end{tabular}
}
\caption{Ablation experiments of PointDC on the ScanNet-v2 validation set.}
\label{tab:ablation}
\end{table*}

\begin{figure*}[h]
  \centering
  \includegraphics[width=0.85\textwidth]{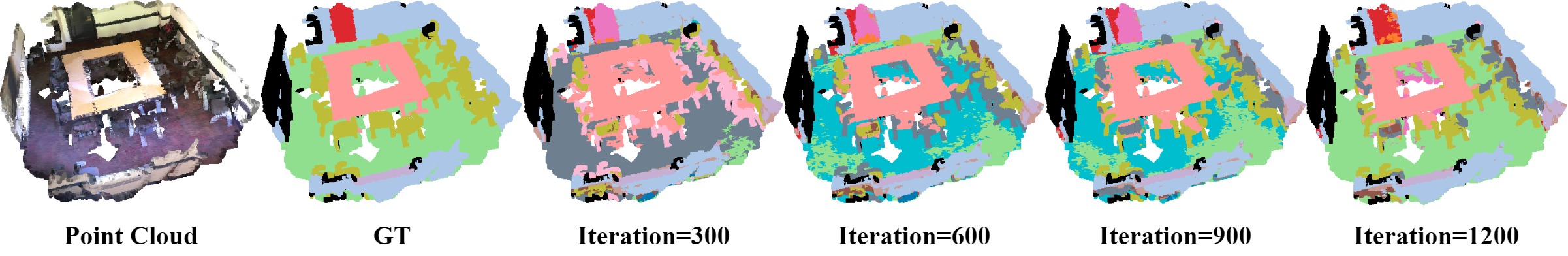}
  \caption{Visualization of PointDC's segmentation results under different iterations during training.}
  \label{fig:ablation_iter_visualization}
\end{figure*}






\subsection{3D Unsupervised Semantic Segmentation}
\label{sec:3d_unsup_segmentation}

\noindent\textbf{Evaluation on ScanNet-v2:}
We conduct the unsupervised clustering and linear probe test on the ScanNet-v2 validation set and report the results in Tab. \ref{tab:scannet_val}.
Previous methods of learning by clustering including DeepCluster \cite{deepcluster}, IIC \cite{iic}, PiCIE \cite{picie} are compared in the table.
Furthermore, we also compare with previous state-of-the-art unsupervised pre-training methods for point clouds including: PointContrast \cite{pointcontrast}, OcCo \cite{occo}, CrossPoint \cite{crosspoint}, CSC \cite{csc} and STRL \cite{strl}.
In the table, PC-HC and PC-NCE respectively represent the PointContrast model trained with Hardest-Contrastive loss and PointInfoNCE loss respectively.
Without using any kinds of human annotations, our method outperforms all other methods as shown in the table.
In particular, PointDC improves by \textbf{+21.10} unsupervised mIoU, \textbf{+40.88} unsupervised accuracy, \textbf{+23.47} linear probe mIoU, and \textbf{27.41} linear probe compared with the next best baseline.
In Tab. \ref{tab:scannet_test}, we present the evaluation results of unsupervised methods on the ScanNet-v2 test set.
Similarly, we can also find a large improvement of \textbf{+18.4} unsupervised mIoU compared with the next best baseline.
Moreover, the qualitative comparisons on the ScanNet-v2 validation set are provided in Fig. \ref{fig:scannet_visualization}.
In this table, we also compare our method with a concurrent work GrowSP \cite{zhang2023growsp} which has good clustering performance, and the results prove that the proposed method achieve better performance.
As the figure reveals, our method is able to precisely locate semantically meaningful objects compared with other unsupervised methods.

\noindent\textbf{Evaluation on S3DIS:}
We also evaluate the proposed method on S3DIS dataset to validate the effectiveness of the proposed method, and show the results in Tab. \ref{tab:s3dis_val}.
The unsupervised semantic segmentation methods of DeepCluster \cite{deepcluster}, IIC \cite{iic}, PiCIE \cite{picie} and the unsupervised pre-training point cloud methods of PointContrast \cite{pointcontrast}, CSC \cite{csc}, STRL \cite{strl} are used for comparison in the table.
As shown in the table, PointDC achieves an improvement of \textbf{+11.50} unsupervised mIoU and \textbf{+16.70} unsupervised Accuracy.

\subsection{Ablation Experiment}

\noindent\textbf{Ablation Study of PointDC Framework:}
To validate the effectiveness of the proposed PointDC framework, we conduct ablation experiments of each related component of the framework, and present the results in Tab. \ref{tab:ablation}.
In the table, `Random Clustering` means the baseline of learning by clustering on 3D point clouds discussed in Sec. \ref{sec:method:prelinminary}.
`Multi-view Clustering` means that a 2D unsupervised learning by clustering framework \cite{STEGO} is conducted on the multi-view images of point clouds, and the 2D segmentation results are fused to construct 3D segmentation results.
It is compared to demonstrate the difference of results existing framework clustered on 2D images and our PointDC framework clustered on 3D point clouds.
`CMD` represents the model trained with Cross-Modal Distillation (Eq. \ref{eq7}), and `Basic SVC` includes a simplified version of Super-Voxel-Clustering,  only utilizing Eq. \ref{eq8} and Eq. \ref{eq12} in the training.
`Non-parametric Classifier` means that we adopt assign the label to each point based on the distance to the cluster centroids instead of a learnable classifier in previous clustering pipelines \cite{picie}, as shown in Eq. \ref{eq9}.
`Super-Voxel Pooling` means that we apply AVG-Pooling towards the point-wise pseudo labels in super-voxels to filter the noise caused by the irregularity of points, as shown in Eq. \ref{eq10}.
From Tab. \ref{tab:ablation}, it can be found that each component of PointDC can improve the performance of unsupervised 3D segmentation effectively.


\noindent\textbf{Ablation Study of Iteration:}
Since our PointDC framework is a learning-by-clustering framework, the model should converge as the clusters optimize in different iterations.
Hence, we conduct experiments of PointDC under different iterations to validate the results.
As shown in Fig. \ref{fig:ablation_iter_visualization}, we visualize the segmentation results of the proposed method under different iterations.
From the figure, we can find that the model becomes better and better along with the training iterations and the update of clustering results.

\subsection{Free-Model} \label{Free-model}

To validate the effectiveness of our method on different models, we additionally employ DGCNN \cite{dgcnn} for 3D unsupervised semantic segmentation. 
With the same backbone of DGCNN, unsupervised methods of OcCo \cite{occo}, CrossPoint \cite{crosspoint}, and STRL \cite{strl} are used for comparison in Tab. \ref{tab:freemodel}.
It demonstrates that our method outperforms previous best results with \textbf{+9.93} unsupervised mIoU and \textbf{+24.84} unsupervised accuracy.

\begin{table}[t] 
\centering
\resizebox{0.6\linewidth}{!}{
\begin{tabular}{l |c|c}     \hline
    Method & mIoU & Acc \\
    \hline \hline
    \cite{occo} OcCo & 3.17 & 19.97 \\
    \cite{crosspoint} CrossPoint & 3.81 & 20.52 \\
    \cite{strl} STRL & 4.13 & 19.38 \\
    \hline 
    PointDC* & 11.50 & 39.65 \\
    PointDC & \textbf{14.06} & \textbf{45.36} \\
    \hline
\end{tabular}
}
\caption{Comparison of unsupervised segmentation methods with the same backbone of DGCNN on ScanNet-v2 validation set. * denotes that only CMD is used.}
\label{tab:freemodel}
\end{table}

\section{Conclusion}

We take the first attempt at the challenging topic of unsupervised semantic segmentation of 3D point clouds without any human annotations, and introduce a novel framework, PointDC.
It contains two steps: Cross-Modal Distillation (CMD) and Super-Voxel Clustering (SVC).
In the first stage of CMD, the multi-view features of the point cloud are back-projected to the 3D space and aggregated together in the super-voxels to distill the training of point representation.
In the next stage of SVC, the point representations are aggregated to super-voxels and then fed to the iterative clustering process for learning semantically meaningful representations.
As the evaluation results on different point cloud benchmarks, our method achieves superior performance on both the ScanNet-v2 (+18.4 mIoU) and S3DIS (+11.5 mIoU).

\section{Acknowledgement}
This work was supported by the National Natural Science Foundation of China (No.61976095) and the Natural Science Foundation of Guangdong Province, China (No.2022A1515010114).
This work was also supported by Alibaba Group through Alibaba Research Intern Program.

{\small
\bibliographystyle{ieee_fullname}
\bibliography{camera_ready}
}

\end{document}